\documentclass{article}
\usepackage{spconf,amsmath,graphicx}

\usepackage{times}
\usepackage{epsfig}
\usepackage{graphicx}
\usepackage{amsmath}
\usepackage{amssymb}
\usepackage{comment}
\usepackage[inline]{enumitem}
\usepackage[caption=false]{subfig}
\usepackage{soul}

\usepackage[utf8]{inputenc} 
\usepackage[T1]{fontenc}    
\usepackage{booktabs}       
\usepackage{amsfonts}       
\usepackage{nicefrac}       
\usepackage{microtype}      
\usepackage{algorithmic}
\usepackage{algorithm}
\usepackage{numprint}
\usepackage{siunitx}
\usepackage{etoolbox}
\newcommand{\ubold}{\fontseries{b}\selectfont}
\robustify\ubold
\usepackage{bbold}

\usepackage{amssymb}
\usepackage{pifont}

\usepackage[pagebackref,breaklinks]{hyperref}
\usepackage{array,multirow,graphicx}

\title{Self-supervised Domain Adaptation in Crowd Counting}

\name{Pha Nguyen$^{1}$, Thanh-Dat Truong$^{1}$, Miaoqing Huang$^{1}$, Yi Liang$^{2}$, Ngan Le$^{1}$, Khoa Luu$^{1}$}
\address{$^{1}$ Department of CSCE, University of Arkansas, Fayetteville, AR, USA \\
$^{2}$ Department of Biological \& Agricultural Engineering, University of Arkansas, Fayetteville, AR, USA \\
\small{\texttt{\{panguyen, tt032, mqhuang, yliang, thile, khoaluu\}@uark.edu}}
}

\newcommand{\xmark}{\ding{55}}%
\newcommand{\cmark}{\ding{51}}%

\begin{document}
%
\maketitle
\begin{abstract}
Self-training crowd counting has not been attentively explored though it is one of the important challenges in computer vision. In practice, the fully supervised methods usually require an intensive resource of manual annotation. In order to address this challenge, this work introduces a new approach to utilize existing datasets with ground truth to produce more robust predictions on unlabeled datasets, named domain adaptation, in crowd counting. While the network is trained with labeled data, samples without labels from the target domain are also added to the training process. In this process, the entropy map is computed and minimized in addition to the adversarial training process designed in parallel.
Experiments on Shanghaitech, UCF\_CC\_50, and UCF-QNRF datasets prove a more generalized improvement of our method over the other state-of-the-arts in the cross-domain setting.
\end{abstract}
\section{Introduction}
\label{sec:intro}

Crowd counting has recently been one of the popular tasks in computer vision. Recent developed methods \cite{wang2020DMCount, wang2021uniformity, song2021rethinking} and datasets \cite{7780439, 6619173, 10.1007/978-3-030-01216-8_33} have been introduced to tackle the counting task with thousands of targets. However, in real-world scenarios, these supervised methods usually learn to count through a training process that requires an extensive annotation of densely populated points in thousands of images. Directly employing models that are trained on existing datasets to a new dataset suffers from a significant performance decrease due to the domain gap.

Therefore, in addition to semantic scene understanding \cite{le2018segmentation} and video temporal modeling \cite{duong2019learning, duong2029cvpr_automatic, Quach_2021_CVPR, Truong_2022_CVPR}, some self-training methods appear to utilize existing datasets with labels, i.e. source domain, and perform counting on more open-set scenarios, i.e. target domain, \cite{8578662,liu2020unsupervised} by transfer learning and domain adaptation techniques. Liu et al. \cite{liu2020unsupervised} enable knowledge distillation between both regression-based and detection-based models by formulating the mutual transformation of outputs. Xu et al. \cite{10.1109ICCV.2019.00847} enhance the generalization over density variance by categorizing image patches into several density levels. While general self learning methods improve the generalization capability by attempting to estimate pseudo ground-truths or distillation learning from a teacher network, a few approaches investigate a new direction to narrow the domain shift from entropy feedback of the target domain, especially in the semantic segmentation task \cite{vu2018advent}.

In this paper, we introduce a new training approach to the crowd counting task toward a domain adaptation setting where the crowd counter utilizes the entropy minimization and adversarial learning to alleviate the distributional discrepancy between the source domain and the target domain.
%
Particularly, our contributions can be summarized as follows:
\begin{itemize}
    \item Reformulate the crowd counting problem from normally estimating density map to directly predicting target points in images, inspired by anchor-based and offset-based approaches.
    
    \item Utilize the Shannon entropy formula as a loss objective function to maximize the prediction certainty.
    
    \item Design an adversarial learning scheme to motivate the network to produce similar distributional predictions over the source domain and the target domain.
    
    \item Evaluate the proposed method with cross-domain settings to demonstrate its substantial generalization compared against the previous crowd counting methods and further perform estimating on a new chicken counting dataset.
\end{itemize}





\begin{figure*}[!t]
    \centering
    \includegraphics[width=0.85\linewidth]{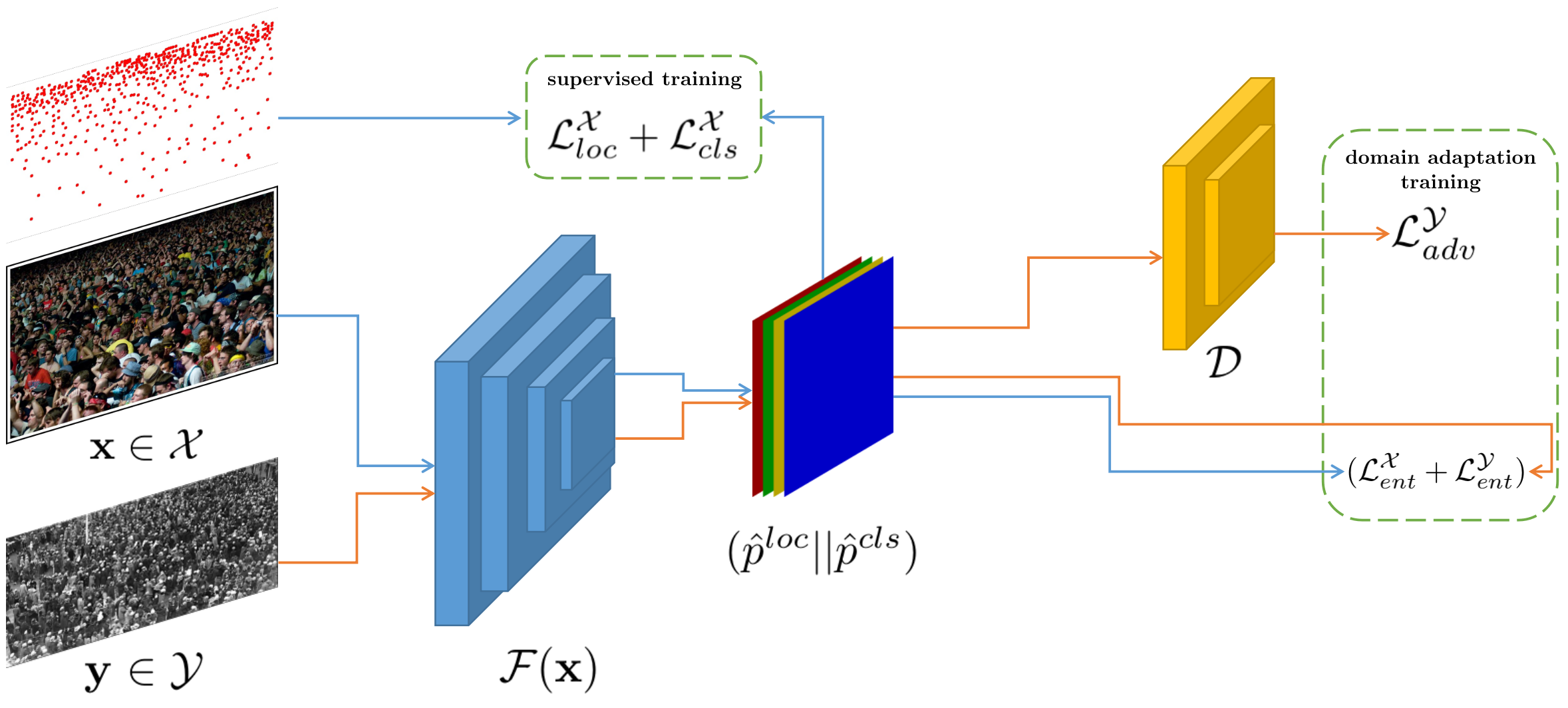}
    \caption{Our overall framework: Given an image sample, the deep network first extracts $\mathcal{F}(\mathbf{x})$ feature, then estimates location offset map and classification map ($\hat{p}^{loc}, \hat{p}^{cls}$). With source domain sample $\mathbf{x} \in \mathcal{X}$, since label is available, supervised L2 Distance $\mathcal{L}_{loc}^{\mathcal{X}}$ loss and Cross Entropy $\mathcal{L}_{cls}^{\mathcal{X}}$ loss can be effortlessly calculated and they are used to guide the network. On the other hand, since sample on target domain $\mathbf{y} \in \mathcal{Y}$ does not have label, $ \mathcal{L}_{ent}^{\mathcal{X}}, \mathcal{L}_{ent}^{\mathcal{Y}}, \mathcal{L}_{adv}^{\mathcal{Y}}$ loss functions are emloyed to additionally teach the domain adaptation learning process. Blue arrows indicate source sample's learning flow, while orange arrows indicate the learning flow of target sample.}\label{framework}
\end{figure*}

\section{Domain adaptation for crowd counting}
\subsection{Point Proposal Network}
Far apart from prior approaches that normally learn to predict a density map \cite{wang2021uniformity, cheng2019learning}, this work designs a network to estimate head points directly. Given an RGB image $\mathbf{x} \in \mathcal{X}$, the training source domain, the deep feature extracted from the backbone network $\mathcal{F}$ can be denoted as $\mathcal{F}(\mathbf{x})$ and its output size is $W\times H\times D$. $\mathcal{F}(\mathbf{x})$ involves a hyper-parameter $s$ that is the backbone's downscale stride. In particular, each cell on the feature map $\mathcal{F}(\mathbf{x})$ basically is correspondence to a window size $s \times s$ on the original input $\mathbf{x}$. The maximum number of points that can exist in the window is $D$ (point's index is denoted as $k, k \in [0, D - 1]$). Then, given the processed feature map $\mathcal{F}(\mathbf{x})$, two network branches are adopted to predict the point coordinate (denoted as $\hat{p}^{loc}$) and background-foreground classification (denoted as $\hat{p}^{cls}$). From the location $(i, j)$ where the pixel is located in the feature map $\mathcal{F}(\mathbf{x})$, the regression branch learns to estimate $2\times k$ offset values $({\delta_i}_k, {\delta_j}_k)$ in the range $[-1,1]$. The point location $\hat{p}^{loc}_{i,j,k} = (\hat{x}_k, \hat{y}_k)$ is computed as follows:
\begin{equation}
    \begin{split}
        \hat{x}_k = s(i + {\delta_i}_k) \\
        \hat{y}_k = s(j + {\delta_j}_k)
    \end{split}
\end{equation}
In the classification task, two predicted scores belong to positive class $pos_{k}$ (object's point) and negative class $neg_{k}$ (background). The Softmax function is employed to normalize two confident scores $\hat{p}^{cls}_{i,j,k} = (\hat{cls}^{pos}_{k}, \hat{cls}^{neg}_{k})$ that follow a probability distribution whose total sums up to one:
\begin{equation}
    \begin{split}
        \hat{cls}^{pos}_{k} = \frac{e^{pos_{k}}}{e^{pos_{k}} + e^{neg_{k}}} \\
        \hat{cls}^{neg}_{k} = \frac{e^{neg_{k}}}{e^{pos_{k}} + e^{neg_{k}}}
    \end{split}
\end{equation}
\textbf{Supervised Training Losses.} On the source domain $\mathcal{X}$ where labels are provided, the supervised training losses on both branches are formulated as the standard ones. 
The $\ell_2$ distance and Cross Entropy losses are adopted for the regression branch and the classification branch, respectively. 
Denoting $p^{loc}_{i}, cls^{pos}_{i}, cls^{neg}_{i}$ as corresponding ground-truth values of $\hat{p}^{loc}_{i}, \hat{cls}^{pos}_{i}, \hat{cls}^{neg}_{i}$, those loss functions are defined as follows:
\begin{equation}
    \mathcal{L}_{loc}(\mathbf{x}) = \frac{1}{|N|}\sum_{i=1}^{|N|}||\hat{p}^{loc}_{i} - p^{loc}_{i}||_2
    \label{location_loss}
\end{equation}
\begin{equation}
    \mathcal{L}_{cls}(\mathbf{x}) = -\frac{1}{|M|}\sum_{i=1}^{|M|}(cls^{pos}_{i}\log\hat{cls}^{pos}_{i} + cls^{neg}_{i}\log\hat{cls}^{neg}_{i})\label{classification_loss}
\end{equation}
where $N$ is the set of points of the ground truth and $M$ is the set of proposals containing both negative and positive pixel points. $M$ can be obtained from a one-to-one matching strategy (i.e. Hungarian algorithm \cite{8827968, 7780624, song2021rethinking}). Finally, the fully supervised training loss can be obtained as follows:
\begin{equation}
    \mathcal{L}_{loc}^{\mathcal{X}} + \mathcal{L}_{cls}^{\mathcal{X}}
\end{equation}
where $\mathcal{L}^{\mathcal{X}}$ denotes a particular loss calculated on all samples from the source domain $\mathcal{X}$.

\subsection{Entropy Minimization on Target Domain}
On the target domain $\mathcal{Y}$, where labels are not available, while some approaches utilize output from a teacher model as a pseudo-label with lower confidence to guide the learning process 
\cite{10.1007/978-3-030-58621-8_13, Liu2020SemiSupervisedCC, Meng_2021_ICCV},
entropy minimization is a more preferable principle in self-training semantic segmentation demonstrated through a number of research works \cite{vu2018advent, pan2020unsupervised, truong2021bimal}. By formulating the point's head classification similar to the semantic segmentation problem, the Shannon entropy formulation \cite{6773024} can be adopted to be a loss function in order to encourage the deep network to produce a higher confidence score. Given an RGB image $\mathbf{y} \in \mathcal{Y}$ on the target domain, the classification per pixel entropy can be formulated as follows:
\begin{equation}
    \mathcal{E(\mathbf{y})}_{i, j, k} = \frac{-1}{\log{2}}(\hat{cls}^{pos}_{k}\log\hat{cls}^{pos}_{k} + \hat{cls}^{neg}_{k}\log\hat{cls}^{neg}_{k})
\end{equation}
And the self-training entropy loss can be defined as:
\begin{equation}
    \mathcal{L}_{ent}(\mathbf{y}) = \frac{1}{W\times H\times D}\sum_{i}^{W}\sum_{j}^{H}\sum_{k}^{D}\mathcal{E}(\mathbf{y})_{i, j, k}\label{entropy_loss}
\end{equation}
\subsection{Distribution Discrepancy Minimization by Adversarial Learning}

To further narrow the domain gap, we utilize a discriminator $\mathcal{D}$, which is a fully convolutional neural network classifier, to motivate the network to extract similar distribution output over both domains. This discriminator tries to determine which domain the input belongs to by learning domain classification  $(\mathcal{D}_{\mathcal{X}}, \mathcal{D}_{\mathcal{Y}})$, while the main network tries to make the discriminator produce fault predictions. Given the concatenation of offset and category maps from the network $(\hat{p}^{loc} || \hat{p}^{cls})$, the loss function of the discriminator can be formulated as follows,
\begin{equation}
    \begin{split}
        \mathcal{L}_{dis}(\hat{p}^{loc} || \hat{p}^{cls}) = -\sum_{i}^{W}\sum_{j}^{H}[(1-z)\log\mathcal{D}_{\mathcal{X}}(\hat{p}^{loc} || \hat{p}^{cls}) + \\ z\log\mathcal{D}_{\mathcal{Y}}(\hat{p}^{loc} || \hat{p}^{cls})]\label{discriminator_loss}
    \end{split}
\end{equation}
where $z = 0$ if $\hat{p} \equiv \mathcal{F}(\mathbf{x})$ or $z = 1$ if $\hat{p} \equiv \mathcal{F}(\mathbf{y})$, which $\mathbf{x} \in \mathcal{X}, \mathbf{y} \in \mathcal{Y}$, and $(.||.)$ is the tensor concatenation operation.

Additionally, to narrow the produced distributions of source domain and the target domain, we add an adversarial loss in the main network's training process:

\begin{equation}
    \mathcal{L}_{adv}(\mathbf{y}) = -\sum_{i}^{W}\sum_{j}^{H}[\log\mathcal{D}_{\mathcal{X}}(\hat{p}_{\mathbf{y}}^{loc} || \hat{p}_{\mathbf{y}}^{cls})]\label{adversarial_loss}
\end{equation}

More specifically, the adversarial loss is designed to maximize the probability of the discriminator predicting source domain class given target domain samples $\mathbf{y} \in \mathcal{Y}$.

To summarize, the learning process of the main point proposal network involves Eqn. \ref{location_loss}, \ref{classification_loss}, \ref{entropy_loss} and \ref{adversarial_loss} loss functions:

\begin{equation}
    \lambda_{loc}\mathcal{L}_{loc}^{\mathcal{X}} + \lambda_{cls}\mathcal{L}_{cls}^{\mathcal{X}} + \lambda_{ent}(\mathcal{L}_{ent}^{\mathcal{X}} + \mathcal{L}_{ent}^{\mathcal{Y}}) + \lambda_{adv}\mathcal{L}_{adv}^{\mathcal{Y}}
\end{equation}
where$\lambda_{loc}$, $\lambda_{cls}$, $\lambda_{ent}$, $\lambda_{adv}$ are weighted parameters to balance corresponding objective functions, $\mathcal{L}^{\mathcal{X}}$ and $\mathcal{L}^{\mathcal{Y}}$ denote particular losses calculated on all samples from domain $\mathcal{X}$ and $\mathcal{Y}$, respectively.
In parallel, the discriminator $\mathcal{D}$ learns with the guidance of Eqn. \ref{discriminator_loss}:
\begin{equation}
    \mathcal{L}_{dis}^{\mathcal{X}} + \mathcal{L}_{dis}^{\mathcal{Y}}
\end{equation}
The entire training procedure is depicted as in Fig. \ref{framework}.

\section{Experimental Results}

\subsection{Ablation Study}

To illustrate the effectiveness of each proposed objective loss in our method, we conduct the ablative experiments as shown in Tab. \ref{losses_table}. 
We slightly add and remove our training strategies on top of the original supervised approach.
The experimental results have shown that our proposed losses have achieved significant improvement.

\begin{table}[!t]
    \centering\caption{Error rates comparison among loss components. Numbers in italic indicate error rates on source domain, while underlined numbers are results on adapted domain.}
    \resizebox{0.8\columnwidth}{!}{\begin{tabular}{|c|c|c|c|c|}
            \hline
            \multirow{2}{*}{Components}                                                        & \multicolumn{2}{c|}{SHTechA} & \multicolumn{2}{c|}{SHTechB}                                                 \\
            \cline{2-5}
                                                                                               & MAE                          & MSE                          & MAE                   & MSE                   \\
            \hline
            \multirow{2}{*}{$\mathcal{L}_{ent}^{\mathcal{X}}$}                                 & \it 54.32                    & \it 90.39                    & \underline{25.36}     & \underline{39.14}     \\
                                                                                               & \underline{162.78}           & \underline{289.47}           & \it 7.92              & \it 11.53             \\
            \cline{1-5}
            \multirow{2}{*}{$\mathcal{L}_{ent}^{\mathcal{Y}}$}                                 & \it 60.76                    & \it 95.34                    & \underline{22.03}     & \underline{34.27}     \\
                                                                                               & \underline{105.48}           & \underline{164.36}           & \it 10.43             & \it 15.60             \\
            \cline{1-5}
            \multirow{2}{*}{$\mathcal{L}_{ent}^{\mathcal{X}}+\mathcal{L}_{ent}^{\mathcal{Y}}$} & \it 54.04                    & \it 89.37                    & \underline{21.58}     & \underline{30.84}     \\
                                                                                               & \underline{87.76}            & \underline{126.53}           & \it 8.03              & \it 11.98             \\
            \cline{1-5}
            \multirow{2}{*}{$\mathcal{L}_{adv}^{\mathcal{Y}}$}                                 & \it 62.83                    & \it 107.42                   & \underline{28.39}     & \underline{47.58}     \\
                                                                                               & \underline{174.59}           & \underline{302.87}           & \it 15.57             & \it 27.38             \\
            \cline{1-5}
            \multirow{2}{*}{$\mathcal{L}_{ent}^{\mathcal{X}}+\mathcal{L}_{ent}^{\mathcal{Y}}+\mathcal{L}_{adv}^{\mathcal{Y}}$}                                                              & \it 57.67                    & \it 93.71                    & \underline{\bf 18.29} & \underline{\bf 26.21} \\
                                                                                               & \underline{\bf 69.21}        & \underline{\bf 95.36}        & \it 8.72              & \it 12.53             \\
            \hline
        \end{tabular}}\label{losses_table}
\end{table}

\subsection{Comparison against SOTA Methods on Public Datasets}

\begin{table}[!t]
    \centering\caption{Error rates comparison between our approach with other domain adaptation (DA) and supervised methods. Numbers in italic indicate error rates on source domain, while underlined numbers are results on adapted domain.}
    \resizebox{0.8\columnwidth}{!}{\begin{tabular}{|c|c|c|c|c|c|}
            \hline
            \multirow{2}{*}{Method}                           & \multirow{2}{*}{DA}     & \multicolumn{2}{c|}{SHTechA} & \multicolumn{2}{c|}{SHTechB}                                                 \\
            \cline{3-6}
                                                              &                         & MAE                          & MSE                          & MAE                   & MSE                   \\
            \hline
            \multirow{2}{*}{DM-Count \cite{wang2020DMCount}}  & \multirow{2}{*}{\xmark} & \it 60.04                    & \it 96.01                    & \underline{22.91}     & \underline{34.69}     \\
                                                              &                         & \underline{142.00}           & \underline{241.02}           & \it 7.33              & \it 11.87             \\
            \hline
            \multirow{2}{*}{UEPNet \cite{wang2021uniformity}} & \multirow{2}{*}{\xmark} & \it 55.26                    & \it 91.94                    & \underline{24.36}     & \underline{37.22}     \\
                                                              &                         & -                            & -                            & \it 6.38              & \it 10.88             \\
            \hline
            \multirow{2}{*}{P2P \cite{song2021rethinking}}    & \multirow{2}{*}{\xmark} & \it 53.02                    & \it 88.48                    & \underline{21.91}     & \underline{33.86}     \\
                                                              &                         & \underline{158.30}           & \underline{267.51}           & \it 6.55              & \it 9.50              \\
            \hline
            \multirow{2}{*}{ConvNets \cite{8578662}}                             & \multirow{2}{*}{\cmark} & \it 73.5                    & \it 112.3                    & \underline{49.1} & \underline{99.2} \\
                                                              &                         & \underline{140.4}        & \underline{226.1}        & \it 18.7              & \it 26.0             \\
            \hline
            \multirow{2}{*}{SPN+L2SM \cite{10.1109ICCV.2019.00847}}                             & \multirow{2}{*}{\cmark} & \it 64.2                    & \it  98.4                    & \underline{21.2} & \underline{38.7} \\
                                                              &                         & \underline{126.8}        & \underline{203.9}        & \it 7.2              & \it 11.1             \\
            \hline
            \multirow{2}{*}{RDBT \cite{liu2020unsupervised}}                             & \multirow{2}{*}{\cmark} & \it -                    & \it -                    & \underline{\bf 13.38} & \underline{29.25} \\
                                                              &                         & \underline{112.24}        & \underline{218.18}        & \it -              & \it -             \\
            \hline
            \multirow{2}{*}{Ours}                             & \multirow{2}{*}{\cmark} & \it 57.67                    & \it 93.71                    & \underline{18.29} & \underline{\bf 26.21} \\
                                                              &                         & \underline{\bf 69.21}        & \underline{\bf 95.36}        & \it 8.72              & \it 12.53             \\
            \hline
        \end{tabular}}\label{SHTech_data}
\end{table}

\textbf{Shanghaitech Dataset} \cite{7780439} consists of two parts: Part-A and Part-B and it contains totally 1,198 images of 330,165 people. We use these two parts to take turns as source and target domains as shown in Tab. \ref{SHTech_data}.  In each method, the first row is using SHTechA for the source domain, SHTechB for the target domain, and the second row is trained in reversed order. The results show that, with domain adaptation learning, our method can be aware of the target's distribution, and yields better quantitative results on its samples (69.21/95.36 vs 112.24/218.18 of RDBT \cite{liu2020unsupervised} on SHTechA), while the performance on source domain is not hurt very much (57.67/93.71 vs 53.02/88.48 on SHTechA and 8.72/12.53 vs 6.55/9.50 on SHTechB of P2P \cite{song2021rethinking}).

\begin{table}[!t]
    \centering\caption{Error rates comparison between our approach with other domain adaptation (DA) and supervised methods.}
    \resizebox{0.9\columnwidth}{!}{\begin{tabular}{|c|c|c|c|c|c|}
            \hline
            \multirow{2}{*}{Method}                           & \multirow{2}{*}{DA}     & \multicolumn{2}{c|}{UCF\_CC\_50} & \multicolumn{2}{c|}{UCF-QNRF}                                                 \\
            \cline{3-6}
                                                              &                         & MAE                          & MSE                          & MAE                   & MSE                   \\
            \hline
            \multirow{1}{*}{DM-Count \cite{wang2020DMCount}}  & \multirow{1}{*}{\xmark} & 427.16                    & 638.92                    & 315.94     & 542.23     \\
            \hline
            \multirow{1}{*}{ConvNets \cite{8578662}}                             & \multirow{1}{*}{\cmark} & 364.0                    & 545.8                    & - & - \\
            \hline
            \multirow{1}{*}{SPN+L2SM \cite{10.1109ICCV.2019.00847}}                             & \multirow{1}{*}{\cmark} & 332.4                    & 425.0                    & 227.2 & 405.2 \\
            \hline
            \multirow{1}{*}{RDBT \cite{liu2020unsupervised}}                             & \multirow{1}{*}{\cmark} & 368.01                    & 518.92                    & 175.02 & 294.76 \\
            \hline
            \multirow{1}{*}{Ours}                             & \multirow{1}{*}{\cmark} & \bf 305.57                    & \bf 400.62                    & \bf 154.73 & \bf 237.84 \\
            \hline
        \end{tabular}}\label{UCF_data}
\end{table}

\noindent
\textbf{UCF\_CC\_50 dataset} \cite{6619173} and \textbf{UCF-QNRF dataset} \cite{10.1007/978-3-030-01216-8_33} have a large variant number of head counts. While the former only contains 50 images but the number of head points varies from 94 to 4,543, the latter consists of 1,535 images with 1,251,642 point heads in total. We use Shanghaitech Part-A for the source domain to adapt on these two datasets. The results also prove our method with domain adaptation perform superior quantitative results on target domain as shown in Tab. \ref{UCF_data} (305.57/400.62 vs 332.4/425.0 of SPN+L2SM \cite{10.1109ICCV.2019.00847} on UCF\_CC\_50) and (154.73/237.84 vs 227.2/405.2 of SPN+L2SM \cite{10.1109ICCV.2019.00847} on UCF-QNRF).

\subsection{Qualitative Result on Chicken Counting}

\begin{figure}[!t]
    \centering
    \includegraphics[width=0.9\linewidth]{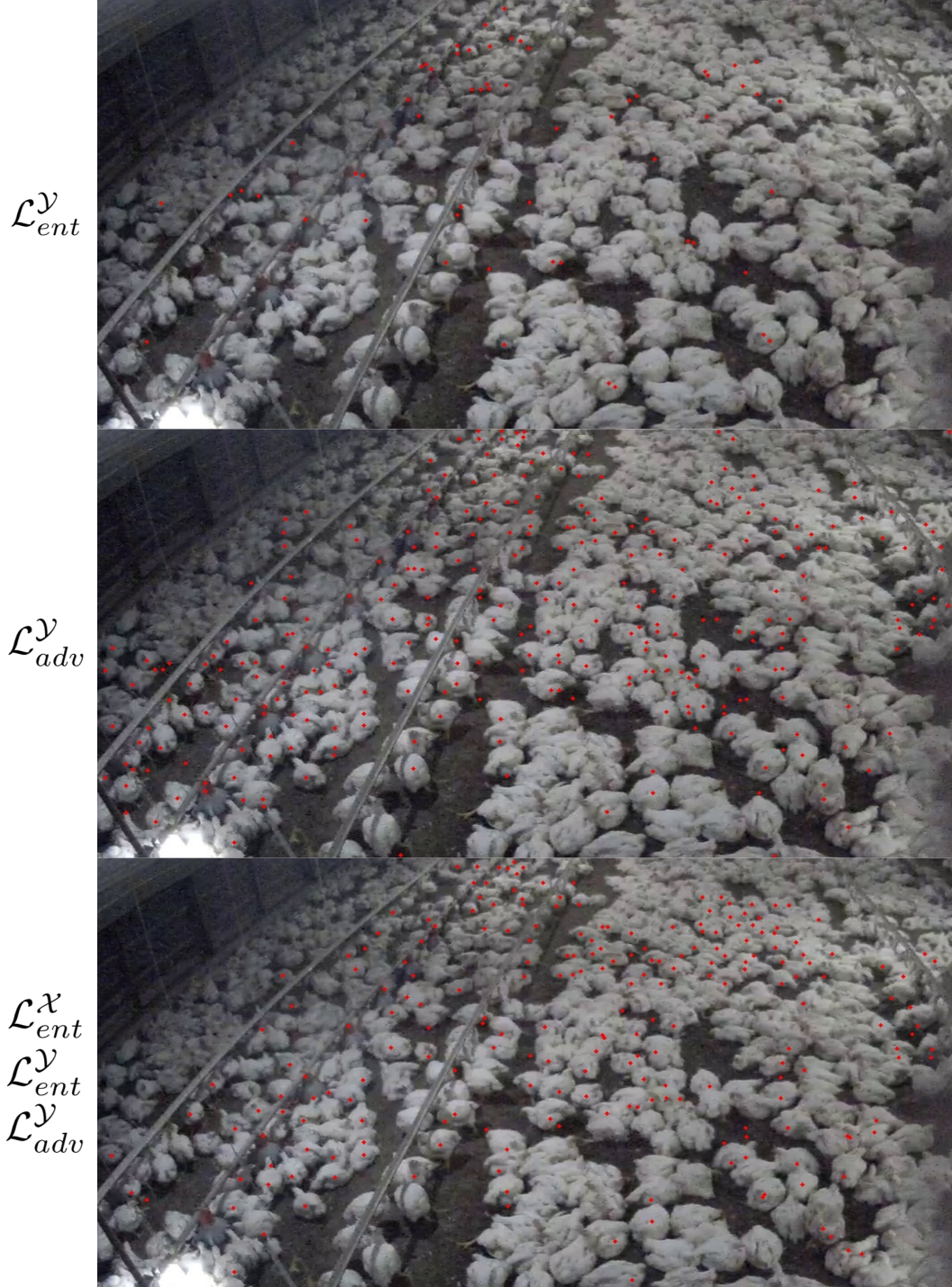}
    \caption{Our qualitative result on our chicken dataset with different domain adaptation training strategies (from top to bottom: entropy minimization loss; adversarial loss; both the losses). Best viewed in color and zoom in.}\label{chicken_dataset}
    \vspace{-6mm}
\end{figure}

We want to evaluate the proposed training method on our chicken dataset collected in farm scenes which have not been annotated as shown in Fig. \ref{chicken_dataset}.
The dataset will be annotated and soon publicly release a test set for quantitative evaluation. We train the SHTech dataset as the source domain and try different domain adaptation training strategies on this dataset.

The first row is the training process with entropy minimization on the target domain. Since the network is mainly guided to learn the localization and classification tasks from the human dataset, the network finds it difficult to recognize chickens as positive class and the result mostly returns false negatives.
The second row is the training process with adversarial loss. While the distribution gap is more narrow resulting in more densely populated prediction, the network produces more false positives by trying to map the dense distribution of the source domain.
The final training process balances those loss functions with weighted parameters and refines better results. However, it still does not yield optimal predictions and there are some missing counts caused by different lighting conditions (i.e. darker and brighter areas in top-left and bottom-left corners).

\section{Conclusion}

In this paper, we have proposed a domain adaptation training scheme for the crowd counting task. Our method is designed to minimize the domain gap between the source domain and the target domain through the entropy loss and the adversarial loss. The entropy minimization is computed on both domains while the adversarial objective minimizes the distribution discrepancy on target samples. As a result, our proposed method shows better results on the target domain than recent self-training learning methods, while maintaining nearly the same error rates on the source domain. Furthermore, we show qualitative estimation on our chicken dataset which is used as the target domain. However, there are still some false negative counts on chickens, due to the lighting condition problem which is not fully addressed in this work. The dataset will be released and the limitation will be studied more in future work.

\textbf{Acknowledgement} This work is supported by NSF Data Science, Data Analytics that are Robust and Trusted (DART) and the Chancellor’s Innovation and Collaboration Fund from University of Arkansas Fayetteville.

\bibliographystyle{IEEEbib}
\small{
    \bibliography{refs}
}
\end{document}